\documentclass{article}

    \PassOptionsToPackage{numbers, compress}{natbib}

\usepackage[preprint]{neurips_2025}

\usepackage[utf8]{inputenc} 
\usepackage[T1]{fontenc}    
\usepackage{hyperref}       
\usepackage{url}            
\usepackage{booktabs}       
\usepackage{amsfonts}       
\usepackage{tcolorbox}      
\usepackage{amsmath}        
\usepackage{nicefrac}       
\usepackage{microtype}      
\usepackage{xcolor}         
\usepackage{graphicx}
\usepackage{subcaption}
\usepackage{xspace}
\usepackage{enumerate}
\usepackage{enumitem}
\usepackage{fdsymbol}
\usepackage{fontawesome}
\usepackage{multirow}
\usepackage{float}

\newcommand{\model}{\texttt{\textsc{BrowseComp-Plus}}\xspace}

\newcommand{\tableformat}{
\setlength{\tabcolsep}{4pt}
\renewcommand{\arraystretch}{1.0}
\small
}

\title{BrowseComp-Plus: A More Fair and Transparent Evaluation  Benchmark of Deep-Research Agent}

\author{
Zijian Chen\thanks{Equal Contribution.\quad \faEnvelopeO ~Correspondence: x93ma@uwaterloo.ca}~~$^{\text{,}1}$,
Xueguang Ma$^{*\text{,~\faEnvelopeO}}$$^{\text{,}1}$,
Shengyao Zhuang$^*$$^{\text{,}2\text{,}5}$,
Ping Nie$^3$,
Kai Zou$^3$,\\
\textbf{
Andrew Liu$^1$,
Joshua Green$^1$,
Kshama Patel$^1$,
Ruoxi Meng$^1$,
Mingyi Su$^1$,}\\
\textbf{
Sahel Sharifymoghaddam$^1$,
Yanxi Li$^1$,
Haoran Hong$^1$,
Xinyu Shi$^1$,
Xuye Liu$^1$,}\\
\textbf{
Nandan Thakur$^1$,
Crystina Zhang$^1$,
Luyu Gao$^4$,
Wenhu Chen$^1$,
Jimmy Lin$^1$}
\\
[1ex]
$^1$University of Waterloo,\quad
$^2$CSIRO,\quad
$^3$Independent,\\
$^4$Carnegie Mellon University,
$^5$The University of Queensland,\quad
\\
[1.5ex]
\url{https://texttron.github.io/BrowseComp-Plus/}
}

\begin{document}

\maketitle

\begin{abstract}
\textbf{Deep-Research} agents, which integrate large language models (LLMs) with search tools, have shown success in improving the effectiveness of handling complex queries that require iterative search planning and reasoning over search results.
Evaluations on current benchmarks like BrowseComp relies on black-box live web search APIs, have notable limitations in (1) \textbf{\textit{fairness}}: dynamic and opaque web APIs hinder fair comparisons and reproducibility of deep research methods; (2) \textbf{\textit{transparency}}: lack of control over the document corpus makes it difficult to isolate retriever contributions. In other words, the current evaluations may compare a complete deep research system at a given time, but they do not foster well-controlled experiments to provide insights into the capability of underlying deep research LLMs.
To address these challenges, we introduce \model, a benchmark derived from BrowseComp, employing a fixed, carefully curated corpus.
Each query in \model includes human-verified supporting documents and mined challenging negatives, enabling controlled experimentation.
The benchmark is shown to be effective in distinguishing the performance of deep research systems.
For instance, the open-source model Search-R1, when paired with the BM25 retriever, achieves 3.86\% accuracy, whereas the GPT-5 achieves 55.9\%.
Integrating the GPT-5 with the Qwen3-Embedding-8B retriever further enhances its accuracy to 70.1\% with fewer search calls.
This benchmark allows comprehensive evaluation and disentangled analysis of deep research agents and retrieval methods, fostering insights into retrieval effectiveness, citation accuracy, and context engineering in Deep-Research system. 

\end{abstract}

\begin{figure}[h]
    \centering
    \includegraphics[width=0.9\textwidth]{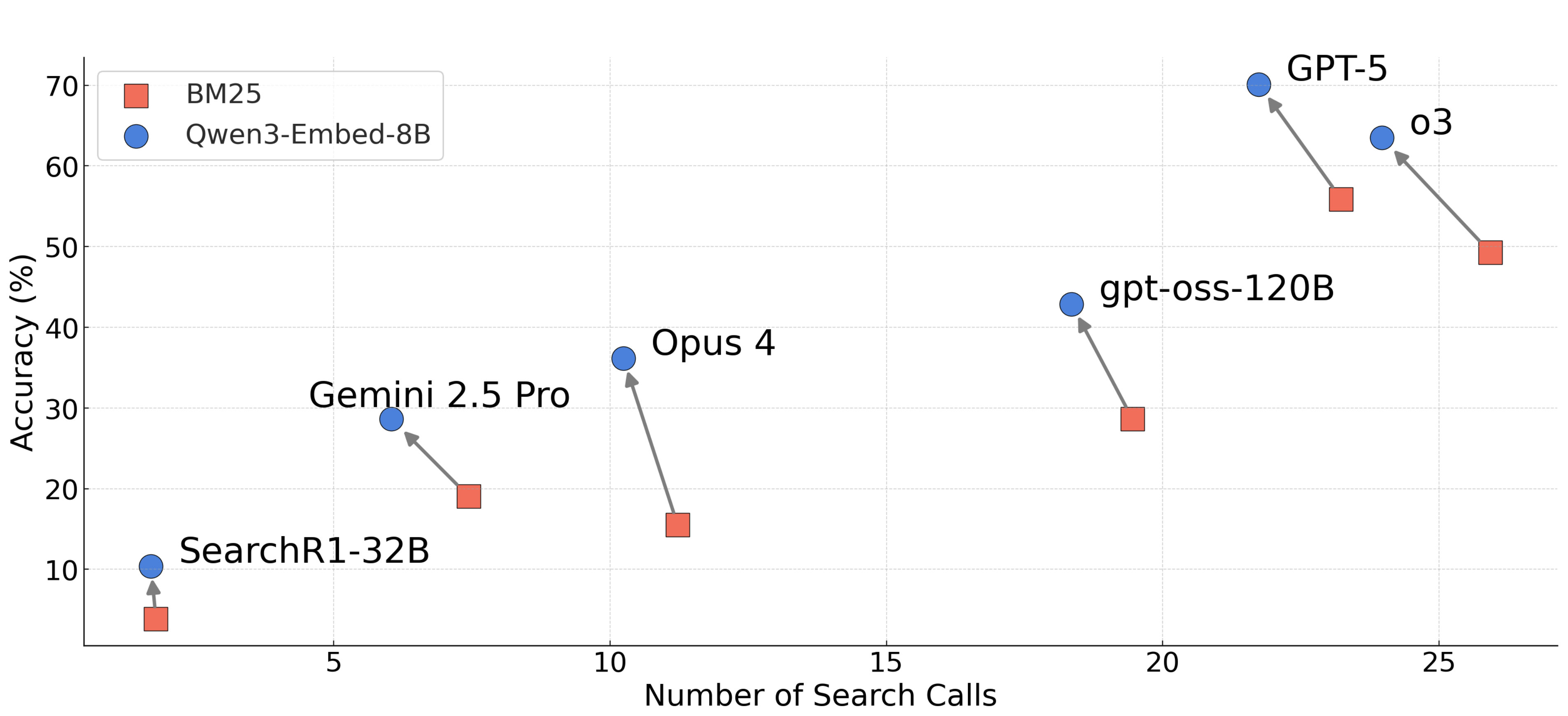}
    \caption{Accuracy vs.\ number of search calls for Deep-Research agents with different retrievers. GPT5, o3, gpt-oss are evaluated with high reasoning effort. 
    The figure shows that \textbf{Deep Research agents mostly improve the final accuracy at a cost of more search calls,
    }
    whereas \textbf{better retrieval systems not only improve the overall accuracy but also reduce the number of search calls}.
    }
    \label{figure:teaser}
\end{figure}

\section{Introduction}
Recent benchmarks for evaluating Deep-Research Agents, such as BrowseComp~\cite{wei2025browsecompsimplechallengingbenchmark}, have showcased the impressive capabilities of combining large language models (LLMs) with web search tools in solving complex, reasoning-intensive queries~\cite{rag,asai2024selfrag}. These benchmarks typically provide sets of queries paired directly with answers, agents are employed with live web search APIs to retrieve supporting documents in real time~\cite{zhou2025browsecompzhbenchmarkingwebbrowsing,chen2025medbrowsecompbenchmarkingmedicaldeep}. While this approach effectively assesses the end-to-end performance of Deep-Research agents, it introduces several critical limitations that impede systematic analysis and evaluation of individual system components.

\begin{itemize}


\item \textbf{{Fair Comparison on Deep Research Agents.}}
{Current evaluations of deep-research agents often conflate agent system performance with the effectiveness of their retrieval components, making it difficult to achieve fair and consistent comparisons across systems.}
{This entanglement also severely undermines} the reproducibility of experiments, which is a key requirement for rigorous evaluation~\cite{Voorhees2019}.

\item 
\textbf{{Transparency of Retrieval Process.}}
{ 
The transparency of the retrieval process comes from two aspects:\
the retrieval algorithm and the target retrieval corpus.
In the current evaluation pipelines, supporting documents are obtained through black-box web search APIs that operate over the entire internet,
which are highly dynamic in content and consistently evolving over time. 
The lack of a controlled retrieval process hinders the evaluation of retrieval models' contribution to deep-research agents.
}

\item \textbf{Accessibility}: The dependence on commercial web search APIs introduces substantial practical constraints, including high operational costs and variability in retrieval quality. These issues not only limit accessibility but also introduce unnecessary complexity and uncertainty in benchmarking.
\end{itemize}

To address these limitations and enable precise, reproducible, transparent, and component-focused evaluation of Deep-Research agents, we introduce \model, a novel benchmark dataset. \model extends the original BrowseComp dataset~\cite{wei2025browsecompsimplechallengingbenchmark} by providing a fixed and curated corpus of documents specifically selected and verified by human annotators.
Each query in \model is accompanied by explicitly identified supportive documents and hard negative documents. This carefully collected document corpus allows researchers to evaluate the retrieval and LLM agent components independently, facilitating detailed analysis of each component’s impact on the final answer quality. Additionally, by eliminating reliance on dynamic web APIs, \model significantly reduces costs, enhances reproducibility, and improves the overall robustness of benchmarking in Deep-Research.

To demonstrate the utility of \model, we conduct comprehensive evaluations by pairing various open- and closed-source LLMs with a range of retrieval models on our curated corpus. This setup allows us to systematically analyze how different combinations affect answer quality and to identify where performance bottlenecks lie, whether in the retriever or the language model. We find that even when equipped with state-of-the-art retrievers, Deep-Research agents still face substantial challenges in consistently surfacing all necessary evidence, for reasoning-intensive queries. These findings motivate the need for evaluation frameworks that disentangle retrieval from reasoning, support fine-grained component analysis, and remain fully reproducible.

Furthermore, we extend our evaluation to test retrieval models directly on the original BrowseComp queries, an analysis that was previously infeasible due to the absence of a fixed corpus and grounded relevant document judgments. Our findings reveal that even state-of-the-art retrieval models struggle to retrieve relevant documents for these complex, reasoning-intensive queries, highlighting a substantial gap in current retrieval capabilities and pointing to important directions for future research in information retrieval.

In summary, our contributions are threefold:
\begin{itemize}
\item We present \model, a fair and transparent benchmark for Deep-Research Agents, featuring a fixed, human-verified corpus with both supporting and challenging negative documents.
\item We provide the first systematic analysis of retrieval–agent interactions under controlled conditions, evaluating a broad range of retrievers and LLM-based agents.
\item We release all benchmark data, evaluation scripts, and baselines to facilitate reproducible research and foster future advances in various dimensions to improve the deep-research system.
\end{itemize}


\section{Related Works}

\subsection{Deep-Research Agent}
Recent advancements in leveraging LLMs for complex query answering have demonstrated the effectiveness of interactions with external retrieval tools. Deep research agents perform tasks with iterative query reasoning, search planning, and reflection on retrieved results~\cite{asai2024selfrag} outperforming the traditional single-round retrieval-agumented generation paradigm~\cite{rag}.
Commercial closed-source models such as Gemini~\cite{gemini}, Opus~\cite{anthropic2024claude3}, and o3~\cite{o3}, and open-source models like GPT-OSS~\cite{openai2025_gptoss} allow access to external retrievers via tool-usage functionality or MCP~\cite{anthropic2024_mcp}. Recent research works like Search R1~\cite{jin2025searchr1trainingllmsreason} and WebSailor~\cite{li2025websailornavigatingsuperhumanreasoning}, built on the Qwen~\cite{yang2025qwen3technicalreport} model, leverage reinforcement learning to further enhance search tool capabilities.

However, fairly evaluating the capabilities of Deep-Research agents requires a fixed retriever system for consistent comparisons. Existing studies mostly evaluate Deep-Research agents using black-box web search APIs. \model addresses this gap and enables fair comparisons across different LLM search agents.

\subsection{Neural Retrieval}
Neural retrieval methods, such as Dense Passage Retrieval~\cite{karpukhin-etal-2020-dense}, encode queries and documents into dense vectors using transformer models and perform retrieval through nearest-neighbor search~\cite{douze2024faiss}.
These methods have significantly improved retrieval effectiveness compared to traditional lexical-based methods like BM25~\cite{robertson1994okapi_trec3}.

Recent improvements in neural retrievers include advanced training strategies such as continuous pretraining~\cite{bge-m3, gao-callan-2022-unsupervised}, data augmentation~\cite{li2023generaltextembeddingsmultistage, ma-etal-2025-drama, shao2025reasonirtrainingretrieversreasoning}, integration of large language models as backbones~\cite{repllama, wang2023improving}, and LLM distillation techniques~\cite{lee2024geckoversatiletextembeddings, zhang2025qwen3embeddingadvancingtext}.
These innovations enhance both effectiveness and generalizability.
While retrievers are a critical component of deep research agents, the contribution of different retrievers to the overall performance of these agents remains underexplored. \model allows systematic evaluation of various neural retrievers as a search tool for Deep-Research agents.

\subsection{Deep Retrieval Benchmarks}
Traditional benchmarks such as NaturalQuestions~\cite{kwiatkowski-etal-2019-natural} and TriviaQA~\cite{joshi-etal-2017-triviaqa} have significantly contributed to evaluating retrieval and retrieval-augmented generation systems~\cite{rag, karpukhin-etal-2020-dense,lin2024radit}.
However, these benchmarks primarily feature single-hop questions, which typically do not require multi-step reasoning or iterative retrieval.
Although datasets like HotpotQA~\cite{yang-etal-2018-hotpotqa} offer multi-hop questions, their corpus is limited to Wikipedia, which is extensively covered during the training of LLMs.

To robustly evaluate deep research systems capable of complex reasoning and strategic search planning, benchmarks requiring sophisticated multi-turn query interactions are essential. BrowseComp~\cite{wei2025browsecompsimplechallengingbenchmark} stands out as a benchmark explicitly designed for this purpose, offering complex queries paired with verifiable answers.
Recent extensions of BrowseComp concepts, such as ZH-BrowseComp~\cite{zhou2025browsecompzhbenchmarkingwebbrowsing} and MedBrowseComp~\cite{chen2025medbrowsecompbenchmarkingmedicaldeep}, further expand to multilingual queries and domain-specific challenges.
Existing benchmarks primarily focus on question-answer evaluations of integrated systems without standardized corpora, complicating comparative assessments of retrieval methodologies. \model facilitates fair and comprehensive evaluations by providing human-verified corpus.
\section{BrowseComp-Plus}
In this section, we provide details on the construction of the proposed \model dataset, which builds upon BrowseComp~\cite{wei2025browsecompsimplechallengingbenchmark} to further enable independent evaluation of the retrieval and LLM components within the Deep-Research framework.

\subsection{Preliminary: BrowseComp}
The BrowseComp benchmark comprises 1,266 challenging fact-seeking questions specifically designed to assess the capability of Deep-Research AI agents to interactively and creatively navigate the web for complex, hard-to-find information~\cite{wei2025browsecompsimplechallengingbenchmark}. The questions are deliberately constructed to be difficult for both humans and LLMs, yet they feature verifiable, concise answers, enabling straightforward evaluation through simple answer matching. While effective and widely employed for end-to-end evaluation of integrated deep research systems, this approach complicates the isolated measurement of retrieval effectiveness within these frameworks.

\subsection{Building the Document Corpus}
\label{section:buildingDocuemntCorpus}

Constructing a corpus for BrowseComp questions is non-trivial. Three key challenges must be addressed:

\begin{enumerate}
    \item \textbf{Comprehensive coverage:} The corpus must provide complete evidence to support the entire reasoning chain required to answer each question.
    \item \textbf{Retrieval difficulty:} It should contain enough distracting negative documents so that search agents and retrievers are challenged in locating the correct evidence.
    \item \textbf{Practical size:} The corpus should be large enough to yield reliable research insights, but avoid too-large computation costs for research purposes.
\end{enumerate}

To meet these criteria, we curate evidence documents through a two-stage pipeline involving automated evidence mining followed by human verification, and perform hard-negative mining via web search to attach challenging, distracting documents to each query. The sections below describe this process in detail and present a 100k-document corpus that effectively supports the study of the Deep Research framework.

\begin{figure}[t]
    \centering
    \includegraphics[width=\columnwidth,clip,trim={20px 10px 25px 10px}]{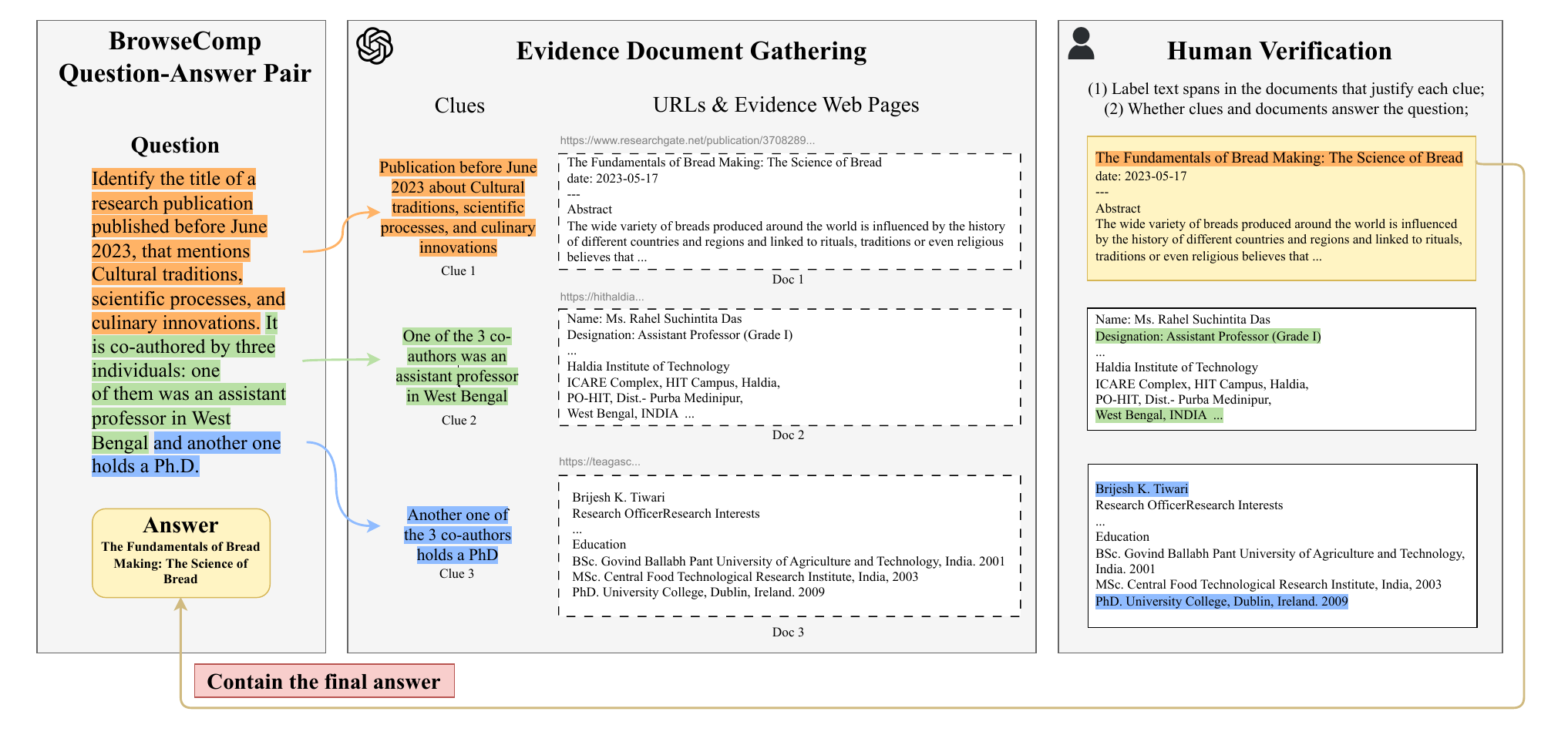}
    \caption{
    The two-stage pipeline of collecting evidence documents in the corpus  (Section~\ref{section:buildingDocuemntCorpus}).
    }
    \label{fig:positive_collection_detailed}
\end{figure}\textbf{}

\subsubsection{Evidence Document Gathering}

The original BrowseComp dataset contains only question-answer pairs, without the URLs of the web pages that support these answers. To build a document collection with supporting evidence, the first step involves retrieving relevant web pages for each question.

To achieve this, we leverage the OpenAI o3 model with web search enabled. We provide the question-answer pairs as input prompts and instruct the model to search online for web pages containing evidence that supports the answers. We also ask the model to structure the output in a table format with three columns: (1) Clue: the part of the question that can help derive the answer; (2) URL: the web page link containing evidence supporting the clue; and (3) Evidence: the content from the web page that supports the clue. The purpose of this table format is to facilitate human annotators in verifying each clue and its corresponding web page in the next step. An example prompt for this step is provided in Appendix~\ref{appendix:gather_doc}.

Of the 1,266 original question-answer pairs in BrowseComp, the OpenAI o3 model fails to provide supporting evidence for 124 pairs, either due to output formatting errors or because the model abstains from answering due to low confidence.
For the remaining 1,142 pairs, we scrape the URLs cited as evidence using Selenium\footnote{\href{https://www.selenium.dev/documentation}{https://www.selenium.dev/documentation}}, and parse them with Trafilatura~\cite{trafilatura}. However, a combination of hallucinated URLs and scraping challenges prevents us from successfully scraping all of them. As a result, we exclude 137 question-answer pairs that contain at least one URL that we are unable to scrape, as missing a URL for a clue will make the question incomplete to answer.

This leaves us with 1,005 queries for the next stage: human verification.

\subsubsection{Evidence Document Verification}
\label{sec:positive_verification}

In this stage, we aim to verify documents that contain evidence for each clue in the questions.
For each question-answer pair, we present human annotators with the output table from OpenAI o3 in the previous stage, with URLs replaced by the corresponding processed documents.

Annotators are asked to:
\begin{enumerate}
    \item [1.] Confirm that each clue is sufficiently justified by the supporting documents. Instead of simply confirming the match, annotators must label the text spans in the documents that justify each clue, as this explicit step encourages high-quality verification.

    \item [2.] Determine whether the combination of clues and supporting evidence enables a human to answer the \emph{entirety} of the question correctly. For instance, if a query asks for an individual matching five characteristics, all five must be verifiable from the documents.
\end{enumerate}

If the original output from OpenAI o3 fails to meet both criteria, annotators are instructed to revise the clues and search the web for additional supporting documents for at least 20 minutes, before concluding that the desired evidence documents cannot be collected.

In addition to constructing the evidence document set, annotators also label which documents directly contain the final answer; these are designated as \emph{gold documents}. Note that a gold document is not defined merely by containing the ground-truth answer as an exact substring; in some cases, the answer is included in the document in an implicit way. For example, a question might ask for the number of publications by a particular author, with the ground-truth answer being ``7''. A gold document in this case could be the author's personal webpage listing their publications; while it may not contain the string ``7'' explicitly, it logically contains the answer. Similarly, there are many cases where the answer appears in the document in a variant form, such as a different date format or a paraphrased phrase, rather than an exact string match. Our goal in constructing the gold document set is to provide a more robust and semantically meaningful alternative to the simple substring-based approach in identifying documents that contain the final answer.

Figure~\ref{fig:positive_collection_detailed} illustrates the complete evidence document collection process. A detailed example, including a screenshot of the labeling interface shown to human annotators, is provided in Appendix~\ref{appendix:ui}.

For quality control, we sample each annotator's labeled data and cross-validate them among annotators, showing over 80\% of agreement on average.
Overall, of the 1,005 question-answer pairs from the previous stage, 830 passed human verification. The most common failure mode occurs when the documents provided by OpenAI o3 do not satisfy the two verification criteria, and human annotators are unable to gather sufficient additional evidence within a reasonable effort.
In addition to these, we identify and exclude several other categories of problematic cases as detailed in Appendix~\ref{sec:problematic}.

The entire labeling process involved 14 university student annotators and required over 400 hours of manual effort.

\subsection{Hard Negative Mining}
\label{section:hardNegativeMining}

\begin{figure}[t]
    \centering
    \includegraphics[width=\columnwidth]{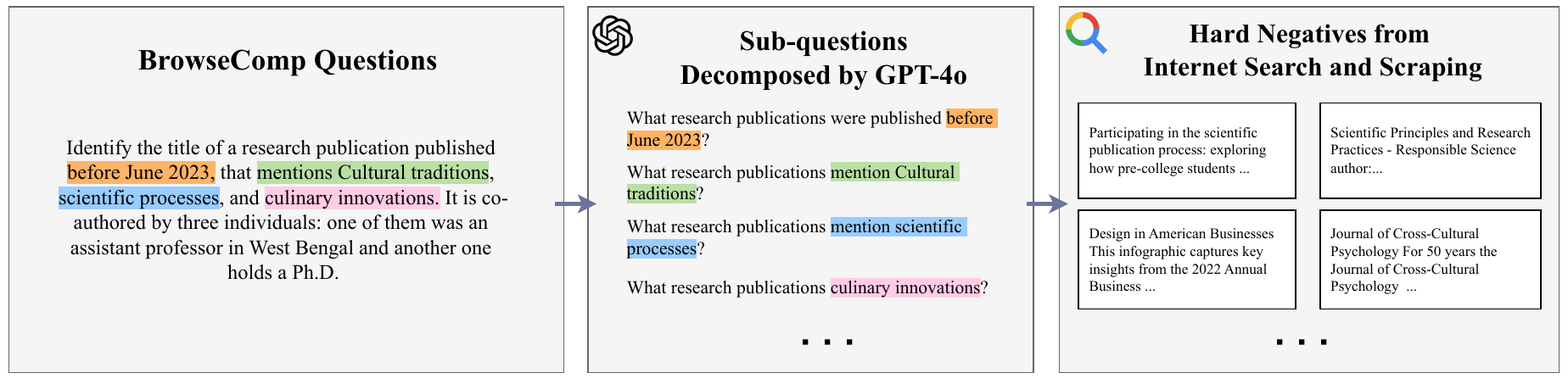}
    \caption{The pipeline of collecting hard negative documents in the corpus(Section~\ref{section:hardNegativeMining}).}
    \label{fig:negative_collection}
\end{figure}


To ensure the collected corpus remains a reasonable size while still being challenging enough for search systems to identify correct answers among distracting documents, we mine hard negative documents via web search to form the corpus. This approach has been proven effective in evaluating information retrieval systems using a small sub-sampled corpus~\cite{corpus_sample,asyncval}.

Specifically, we take each question from BrowseComp and prompt GPT-4o to break it down into simpler, self-contained sub-queries. On average, this results in about seven sub-queries per original query. Each sub-query is then sent to a Google Search API provider (SerpAPI), which returns up to 100 search results. We scrape these results using the same process used for collecting documents during positive example construction. We illustrate this hard negative document collecting process in Figure~\ref{fig:negative_collection}. The prompt used to create these sub-queries is provided in Appendix~\ref{appendix:decompose_prompt}.

\subsection{Final Corpus Statistics}

After deduplicating the positive and negative documents collected as above, we arrive at a corpus of 100,195 documents, along with 830 queries. On average, each query contains 6.1 evidence documents, 76.28 negatives, and 2.9 gold documents. Each document averages 5179.2 words and 32296.2 characters.


\begin{figure}[t]
    \centering
    \includegraphics[width=\columnwidth]{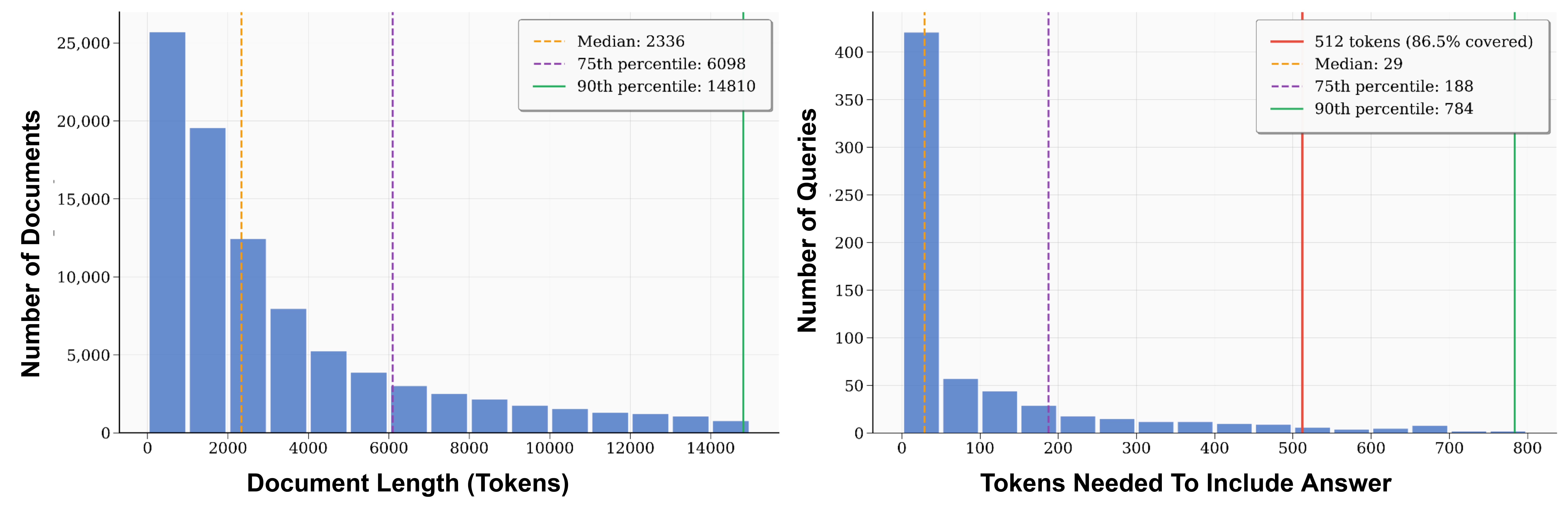}
    \caption{(a) Token distribution of corpus length, showing up to 90th percentile for display; (b) Distribution of tokens needed to include answer in gold documents per query, showing up to 90th percentile for display}
    \label{figure:token_length}
\end{figure}

\section{Experiments}
\label{experiments}

\subsection{Baselines: LLM Search Agents}

We evaluate several representative commercial models with strong agentic search capabilities, ranging from the most advanced reasoning models to cost-effective ones:
o3,
gpt-4.1~\cite{o3},
gpt-5,
claude-opus-4,
claude-sonnet-4~\cite{anthropic2024claude3},
gemini-2.5-pro,
gemini-2.5-flash~\cite{gemini}.

We also assess leading open-source efforts. 
This includes Qwen3-32B~\cite{yang2025qwen3technicalreport}, a popular open-source reasoning LLM, and Search-R1~\cite{jin2025searchr1trainingllmsreason, jin2025empiricalstudyreinforcementlearning}, a model fine-tuned for agentic search based on the Qwen backbone. 
Specifically, we use the 32B checkpoint released in~\cite{jin2025empiricalstudyreinforcementlearning}. 
Finally, we evaluate the recent advanced \texttt{gpt-oss} 20B and 120B~\cite{openai2025_gptoss}, which are reasoning LLMs optimized for search tool usage and offer multiple reasoning effort settings, ranging from low to high.


\subsection{Baselines: Retriever}


In our study, we compared a range of retrieval methods from a traditional lexical baseline to modern state‑of‑the‑art dense embedding retrievers:
\begin{itemize}
\item BM25~\cite{bm25}: The classic sparse lexical retriever, which matches queries to documents based on term statistics.

\item Qwen3-Embedding~\cite{zhang2025qwen3embeddingadvancingtext}: A dense embedding retriever, available in sizes 0.6B, 4B, and 8B, built on the Qwen3 foundation model family~\cite{yang2025qwen3technicalreport}. It achieves state‑of‑the‑art performance on retrieval benchmarks such as MTEB~\cite{muennighoff-etal-2023-mteb}.

\item ReasonIR~\cite{shao2025reasonirtrainingretrieversreasoning}: A dense embedding specifically trained for reasoning‑intensive retrieval via synthetic data generation, setting a new state‑of‑the‑art on reasoning-intensive information retrieval benchmark BRIGHT~\cite{su2025bright}.
\end{itemize}

We use the Pyserini IR toolkit~\cite{pyserini} to serve the BM25 retriever, and the Tevatron dense retrieval toolkit~\cite{tevatron} to serve Qwen3-Embedding and ReasonIR.

\subsection{Experiment Setup}
\paragraph{Search Agents}
To perform agentic search with the LLMs, we provide the LLM with a retriever tool as tool use.
We follow the original prompt from BrowseComp~\cite{wei2025browsecompsimplechallengingbenchmark}, which instructs the model to answer a given question along with a confidence estimate (expressed as a percentage).
There are two revisions of the original prompts: (1) We explicitly prompt the LLM to use the provided tools to adapt our custom search tool; (2) We instruct the model to cite the sources when generating the final answer, enabling the evaluation of citation quality.
The complete prompt is included in Appendix~\ref{appendix:main_prompt}.
We use this prompt across all models except Search-R1, which uses the prompt aligned with its original fine-tuning.

\paragraph{Retriever}

The retriever tool is set to retrieve the top $k = 5$ search results, where each result is truncated to the first 512 token of the corresponding document. This truncation is due to budget constraints, which prevent us from providing full document content. To assess the impact of this design choice, we analyze the distribution of the number of tokens required to include the ground-truth answer for each query. As illustrated in Figure~\ref{figure:token_length} (b), when documents are truncated to the first 512 tokens, 86.5\% of queries still contain the ground-truth answer in at least one of their gold documents.
Further ablations exploring alternative tool configurations are discussed in Section~\ref{sec:getdoc}.

\subsection{Evaluation Metrics}

\paragraph{Deep Research Effectiveness}

We report end-to-end effectiveness of the deep research systems with four metrics: Accuracy, Recall, and Search Calls. 
Accuracy follows BrowseComp: an LLM-as-judge (gpt-4.1) compares the model’s final answer against the ground truth using the evaluation prompt listed in Appendix~\ref{appendix:eval_prompt}.
Recall measures how many human-verified evidence documents the agent retrieved during its entire interaction.
Search Calls is the average number of search API invocations per query.
In addition, following BrowseComp, we compute calibration error using the confidence estimates produced by the search agents, in the same way as Humanity's Last Exam~\cite{phan2025humanitysexam}.
It measures how closely a model's predicted confidence matches the actual accuracy of its predictions.
For Search-R1, we do not report calibration error because the input and output format of this model are fixed without a confidence source output.

\paragraph{Retrieval Effectiveness}
For evaluating retriever effectiveness, our \model benchmark provides human-verified evidence documents and gold documents, along with a fixed test document collection, enabling evaluation under the Cranfield paradigm~\cite{Voorhees2019}.
Specifically, we follow standard TREC practice to create a query-document relevance label file\footnote{Known as a qrel file.} for both evidence documents and gold documents separately, and then compute Recall@k and nDCG@k to assess the effectiveness of retrievers.


\subsection{Results}

\begin{table}[]
  \centering
\caption{End-to-end agent accuracy on \model across LLMs and retrievers. All agents are prompted with the same tool-use prompt, except for Search-R1, which uses the prompt identical to the training.}

  \label{table:main}
  \resizebox{\textwidth}{!}{
    \begin{tabular}{llrrrr}
      \toprule
      \textbf{LLM} & \textbf{Retriever} & \textbf{Accuracy} & \textbf{Recall} & \textbf{Search Calls} & \textbf{Calibration Error} \\
      \midrule
      \multirow{2}{*}{gpt-4.1}
        & BM25                   & 14.58\% & 16.42\% & 10.35 & 68.96\% \\
        & Qwen3-Embed-8B         & 35.42\% & 36.89\% &  8.67 & 54.67\% \\
      \midrule
      \multirow{2}{*}{o3}
        & BM25                   & 49.28\% & 56.64\% & 25.93 & 12.58\% \\
        & Qwen3-Embed-8B         & 63.49\% & 73.24\% & 23.97 & 16.77\% \\
        \midrule
      \multirow{2}{*}{gpt-5}
        & BM25                   & 55.90\% & 61.70\% & 23.23 & 13.50\% \\
        & Qwen3-Embed-8B         & 70.12\% & 78.98\% & 21.74 &  9.11\% \\
      \midrule
      \multirow{2}{*}{Sonnet\,4}
        & BM25                   & 14.34\% & 21.31\% &  9.95 & 29.79\% \\
        & Qwen3-Embed-8B         & 36.75\% & 47.33\% &  9.03 & 24.51\% \\
      \midrule
      \multirow{2}{*}{Opus\,4}
        & BM25                   & 15.54\% & 22.96\% & 11.22 & 22.00\% \\
        & Qwen3-Embed-8B         & 36.14\% & 50.84\% & 10.24 & 12.79\% \\
      \midrule
      \multirow{2}{*}{Gemini~2.5~Flash}
        & BM25                   & 15.54\% & 21.45\% & 10.56 & 29.28\% \\
        & Qwen3-Embed-8B         & 33.01\% & 40.19\% &  9.77 & 21.63\% \\
      \midrule
      \multirow{2}{*}{Gemini~2.5~Pro}
        & BM25                   & 19.04\% & 22.81\% &  7.44 & 51.58\% \\
        & Qwen3-Embed-8B         & 28.67\% & 35.31\% &  6.04 & 44.08\% \\
      \midrule
        \multirow{2}{*}{gpt-oss-120B-high}
        & BM25                   & 28.67\% &35.50\% & 19.45 & 46.48\% \\
        & Qwen3-Embed-8B         & 42.89\% & 52.63\% & 18.35 & 40.34\% \\
      \midrule
      \multirow{5}{*}{Qwen3-32B}
        & BM25                   &  3.49\% &  3.12\% & 0.92 & 57.41\% \\
        & Qwen3-Embed-0.6B       &  4.10\% &  3.45\% & 0.91 & 60.71\% \\
        & Qwen3-Embed-4B         &  7.83\% &  6.20\% & 0.89 & 61.06\% \\
        & Qwen3-Embed-8B         & 10.36\% &  7.80\% & 0.94 & 59.84\% \\
        & ReasonIR               &  9.16\% &  7.59\% & 0.91 & 55.15\% \\
      \midrule
      \multirow{5}{*}{SearchR1-32B}
        & BM25                   &  3.86\% &  2.61\% & 1.78 & \textit{N/A} \\
        & Qwen3-Embed-0.6B       &  5.66\% &  5.30\% & 1.73 & \textit{N/A} \\
        & Qwen3-Embed-4B         &  9.40\% &  7.90\% & 1.68 & \textit{N/A} \\
        & Qwen3-Embed-8B         & 10.36\% & 10.17\% & 1.69 & \textit{N/A} \\
        & ReasonIR               &  9.43\% &  8.37\% & 1.74 & \textit{N/A} \\
      \bottomrule
    \end{tabular}
  }
\end{table}

We report both the end-to-end performance of Deep-Research agents (Table~\ref{table:main}) and the standalone retrieval effectiveness (Table~\ref{table:retrieval}) on the \model benchmark. Our key findings highlight the intricate interplay between retrieval quality, reasoning capability, and agent search behavior.

\subsection{End-to-End Deep-Research Performance}
Table~\ref{table:main} summarizes the overall Deep-Research Performance across different LLMs and retrievers.  
Proprietary models (gpt-4.1, o3, gpt-5, Sonnet-4, Opus-4, Gemini) demonstrate high answer accuracy, with OpenAI’s gpt-5 achieving the highest accuracy (70.12\%) when paired with the Qwen3-Embedding-8B retriever.
Open-source models such as Qwen3-32B and SearchR1-32B lag significantly behind proprietary models. With Qwen3-Embedding-8B as the retriever, Qwen3-32B achieves only 10.36\% accuracy, compared to 35.42\% for gpt-4.1 and 63.49\% for o3. Notably, the only high-performing open-source model we studied is gpt-oss-120B in its high reasoning mode, which achieves 42.89\% accuracy, surpassing Opus 4 when both are paired with Qwen3-Embedding-8B.

In general, closed-source agents call the search tool more frequently than open-source models. For instance, OpenAI’s gpt-5 and o3 issue an average of more than 20 search calls per query, while Qwen3-32B and SearchR1-32B make fewer than 2, despite being explicitly prompted to use the tool. This reflects a test-time scaling effect: more exhaustive search correlates with better outcomes and aligns with prior findings that reasoning-intensive queries benefit from multi-turn, exploratory retrieval.

These results illustrate current limitations in the interleaved reasoning and tool-use capabilities of open-source LLMs, despite their comparable performance when directly given relevant documents (as shown in Section~\ref{sec:oracle}).

\begin{table}[]
  \centering
  \caption{Effectiveness of retrievers. The complete question is used as the query for all retrieval methods for fair comparison.}
\label{table:retrieval}
  \begingroup
  \tableformat
    \begin{tabular}{lrrrr}
      \toprule
      \textbf{Retriever} & \textbf{Recall@5} & \textbf{Recall@100} & \textbf{Recall@1000} & \textbf{nDCG@10} \\
      \midrule
      \multicolumn{5}{c}{\textbf{Evidence Document Retrieval}} \\
      BM25           & 1.2 & 4.7 & 13.7 & 1.6 \\
      Qwen3-Embed-0.6B     & 6.2 & 26.5 & 59.7 & 8.0 \\
      Qwen3-Embed-4B       & 9.8 & 40.2 & 71.8 & 14.0 \\
      Qwen3-Embed-8B       & 14.5 & 47.7 & 76.7 & 20.3 \\
      ReasonIR-8B    & 12.2 & 43.6 & 73.9 & 16.8 \\
      \midrule
      \multicolumn{5}{c}{\textbf{Gold Document Retrieval}} \\
      BM25           & 1.4 & 6.1 & 17.3 & 1.7 \\
      Qwen3-Embed-0.6B     & 8.5 & 30.5 & 66.2 & 7.4 \\
      Qwen3-Embed-4B       & 13.0 & 47.3 & 77.0 & 13.6 \\
      Qwen3-Embed-8B       & 18.5 & 55.8 & 83.5 & 19.5 \\
      ReasonIR-8B    & 15.3 & 49.7 & 78.9 & 15.5 \\
      \bottomrule
    \end{tabular}
  \endgroup
\end{table}

\subsection{Effect of Retrieval Quality}
A consistent trend observed across all models is that stronger retrieval leads to higher final accuracy.

First, consider the retriever's effectiveness on our dataset. We evaluate retrieval performance using the original BrowseComp queries, with results shown in Table~\ref{table:retrieval}. Compared to BM25, Qwen3-Embedding-8B and ReasonIR-8B achieve substantially higher recall and nDCG for both evidence document retrieval and gold document retrieval. Notably, we observe a model size scaling law within the Qwen3 embedding family; larger models consistently perform better, with Qwen3-8B surpassing ReasonIR-8B at the 8B scale.

Now, as indicated in Table~\ref{table:main}, replacing the BM25 retriever with a stronger retriever leads to significant accuracy gains across all LLM agents. For instance, OpenAI’s gpt-5 accuracy improves from 55.9\% to 70.12\%, while Sonnet 4 and Opus 4 both more than double their accuracy. This suggests a strong positive correlation between retrieval effectiveness and research agent accuracy.

Moreover, stronger retrievers potentially reduce the number of search calls. For most proprietary models, Qwen3-Embedding-8B reduces search calls by approximately 1–3 compared to BM25. This shows that better retrieval not only improves effectiveness (accuracy) but also efficiency (fewer tool calls). In Appendix~\ref{sec:api_cost}, we also report differences in proprietary agent API cost when using different retrievers. Agents using Qwen3-Embedding-8B incur lower costs due to fewer input and output tokens, further supporting the efficiency gains enabled by stronger retrieval.

These results are likely due to the higher precision of early search results, which reduces the need for follow-up queries. This is supported by the Recall metric in Table~\ref{table:main}, where stronger retrievers yield higher recall for retrieved documents. In addition, Table~\ref{table:citation} reports the coverage, average number, precision, and recall of the document citations attributed by the agent during answer generation. As the results show, although agents using BM25 issue more search calls, nearly all metrics are lower than those achieved with Qwen3-Embedding-8B. This indicates that documents returned by BM25 are less useful in the iterative deep research process, whereas Qwen3-Embedding-8B provides more relevant and informative documents.

\begin{table}[]
  \centering
\caption{Per-query averages of citation coverage, citation count, precision, and recall for labeled evidence documents. Search-R1 is excluded because its fine-tuned outputs do not contain citations.}

  \label{table:citation}
  \begingroup
    \tableformat
    \begin{tabular}{llcccc}
      \toprule
      \textbf{LLM} & \textbf{Retriever} & \textbf{Coverage} & \textbf{Avg \# Citations} & \textbf{Precision} & \textbf{Recall} \\
      \midrule
      \multirow{2}{*}{gpt-4.1}
        & BM25                   & 57.0\%  & 1.92 & 37.0\% & 16.1\% \\
        & Qwen3-Embedding-8B     & 79.2\%  & 2.54 & 58.5\% & 28.2\% \\
      \midrule
      \multirow{2}{*}{o3}
        & BM25                   & 63.5\%  & 3.27 & 86.7\% & 51.0\% \\
        & Qwen3-Embedding-8B     & 78.0\%  & 3.51 & 91.8\% & 56.2\% \\
      \midrule
      \multirow{2}{*}{gpt-5}
        & BM25                   & 94.9\%  & 3.89 & 71.8\% & 51.3\% \\
        & Qwen3-Embedding-8B     & 98.0\%  & 4.28 & 83.4\% & 62.3\% \\
      \midrule
      \multirow{2}{*}{Sonnet\,4}
        & BM25                   & 76.1\%  & 3.19 & 31.9\% & 21.3\% \\
        & Qwen3-Embedding-8B     & 90.7\%  & 4.19 & 52.4\% & 39.9\% \\
      \midrule
      \multirow{2}{*}{Opus\,4}
        & BM25                   & 74.9\%  & 3.03 & 35.1\% & 22.3\% \\
        & Qwen3-Embedding-8B     & 86.1\%  & 3.82 & 58.9\% & 42.6\% \\
      \midrule
      \multirow{2}{*}{Gemini~2.5~Flash}
        & BM25                   & 74.2\%  & 4.89 & 34.2\% & 21.7\% \\
        & Qwen3-Embedding-8B     & 89.2\%  & 4.75 & 51.5\% & 35.1\% \\
      \midrule
      \multirow{2}{*}{Gemini~2.5~Pro}
        & BM25                   & 53.9\%  & 3.03 & 52.1\% & 31.4\% \\
        & Qwen3-Embedding-8B     & 59.4\%  & 3.49 & 64.9\% & 41.5\% \\
      \midrule
      \multirow{2}{*}{gpt-oss-120B-high}
        & BM25                   & 62.5\% & 3.55 & 50.8\% & 31.5\% \\
        & Qwen3-Embedding-8B     & 76.9\%  & 3.88 & 60.8\% & 38.2\% \\
      \midrule
      \multirow{5}{*}{Qwen3-32B}
        & BM25                   & 87.0\%  & 1.85 &  8.9\% &  2.6\% \\
        & Qwen3-Embedding-0.6B   & 90.1\%  & 1.79 &  8.7\% &  2.5\% \\
        & Qwen3-Embedding-4B     & 91.7\%  & 1.84 & 16.1\% &  4.9\% \\
        & Qwen3-Embedding-8B     & 90.2\%  & 1.78 & 20.0\% &  6.6\% \\
        & ReasonIR               & 95.8\%  & 1.74 & 18.0\% &  5.7\% \\
      \bottomrule
    \end{tabular}
  \endgroup
\end{table}

\subsection{Analysis and Ablation}

\subsubsection{Oracle Retrieval}
\label{sec:oracle}

In addition to comparing progressively stronger retrievers, we also evaluate effectiveness in an extreme oracle setting, where search agents are prompted with all labeled positive documents to answer the questions. In this setup, \texttt{gpt-4.1} achieves an accuracy of 93.49\%. This highlights two key points. First, it showcases the importance of the retriever: if the retriever is of perfect quality, search agents can attain substantially high accuracy on complex reasoning tasks in \model, in contrast to the 14.58\% baseline accuracy of \texttt{gpt-4.1} when using BM25 as the retriever. Second, it validates the quality of the \model corpus itself: \texttt{gpt-4.1}, a non-reasoning model, is able to correctly answer 93.49\% of questions using only the  evidence documents in the corpus. For the remaining 6.51\% of cases, human annotators reviewed each instance and confirmed that the answers are indeed answerable from the positive documents; the errors stem solely from \texttt{gpt-4.1}'s failure to reason correctly.

A similar evaluation with \texttt{Qwen3-32B} yields an accuracy of 83.25\% in the oracle setting; among its errors, 50 (6\%) result from the positive documents exceeding the model's context window. The effectiveness gap between \texttt{Qwen3-32B} and \texttt{gpt-4.1} in this setting is notably smaller than the gap observed in the non-oracle setting. This suggests that open-source models do not substantially lag behind proprietary models in their ability to answer questions when provided with sufficient evidence. Instead, their primary limitation lies in performing interleaved reasoning with the search tool, causing the bigger effectiveness gap observed in Table~\ref{table:main}.

\subsubsection{Impact of Reasoning Effort}
We evaluate how the reasoning effort of LLMs influences answer quality and retrieval behavior. 
To isolate this effect, we focus on the \texttt{gpt-oss} family, which offers three reasoning modes: \emph{low}, \emph{medium}, and \emph{high}. 
These modes differ in the amount of computational effort and deliberation the model applies before producing an answer, with higher modes generally involving longer intermediate reasoning steps. We report results in Table~\ref{table:oss}.

Overall, increasing the reasoning effort leads to substantial improvements in both accuracy and recall for all model sizes and retrievers. 
For example, \texttt{oss-20b} with Qwen3-Embed-8B improves accuracy from 13.37\% in \emph{low} mode to 34.58\% in \emph{high} mode, accompanied by a recall jump from 17.37\% to 49.29\%. 
Similarly, \texttt{oss-120b} with Qwen3-Embed-8B rises from 24.94\% to 42.89\% accuracy across the same progression. 
These gains, however, come with a trade-off: higher reasoning modes dramatically increase the average number of search calls (e.g., from $\approx$2 to $\approx$24 for \texttt{oss-20b} with Qwen3-Embed-8B), implying higher computational and latency costs.

Interestingly, calibration error tends to decrease with higher reasoning effort, suggesting that the models become more aligned between confidence and correctness as they reason more extensively. 
Qwen3-Embed-8B consistently outperforms BM25 across all reasoning settings, highlighting the importance of retriever choice alongside reasoning depth.

These findings indicate that increasing reasoning effort can significantly boost answer quality, but at the cost of retrieval overhead, an important consideration when balancing accuracy and efficiency in deep-research generation systems.

\begin{table}[]
  \centering
  \caption{OpenAI gpt-oss models in different reasoning effort settings}
  \label{table:oss}
  \resizebox{\textwidth}{!}{
    \begin{tabular}{llrrrr}
      \toprule
      \textbf{LLM} & \textbf{Retriever} & \textbf{Accuracy} & \textbf{Recall} & \textbf{Search Calls} & \textbf{Calibration Error} \\
      \midrule
      \multirow{2}{*}{oss-20B-low}
        & BM25                   &  4.11\% &  5.36\% &  1.89 & 40.89\% \\
        & Qwen3-Embed-8B         & 13.37\% & 17.37\% &  1.87 & 36.34\% \\
      \midrule
      \multirow{2}{*}{oss-20B-medium}
        & BM25                   & 16.39\% & 21.96\% & 13.72 & 41.78\% \\
        & Qwen3-Embed-8B         & 29.88\% & 41.31\% & 13.64 & 35.99\% \\
      \midrule
      \multirow{2}{*}{oss-20B-high}
        & BM25                   & 21.08\% & 31.98\% & 26.87 & 33.42\% \\
        & Qwen3-Embed-8B         & 34.58\% & 49.29\% & 23.87 & 27.81\% \\
      \midrule
      \multirow{2}{*}{oss-120B-low}
        & BM25                   &  9.52\% &  8.54\% &  2.06 & 43.59\% \\
        & Qwen3-Embed-8B         & 24.94\% & 22.50\% &  2.21 & 40.96\% \\
      \midrule
      \multirow{2}{*}{oss-120B-medium}
        & BM25                   & 23.73\% & 27.02\% &  9.73 & 45.78\% \\
        & Qwen3-Embed-8B         & 37.59\% & 43.45\% &  9.64 & 41.77\% \\
      \midrule
      \multirow{2}{*}{oss-120B-high}
        & BM25                  & 28.67\% &35.50\% & 19.45 & 46.48\%\\
        & Qwen3-Embed-8B         & 42.89\% & 52.63\% & 18.35 & 40.34\% \\
      \bottomrule
    \end{tabular}
  }
\end{table}


\begin{table}[ht]
  \centering
  \caption{Comparison of Qwen3-32B and gpt-4.1 with and without get-document tool, using Qwen3-Embedding-8B as retriever.}
  \label{table:get_doc}
  \begingroup
    \tableformat
    \begin{tabular}{lcccc}
      \toprule
      \textbf{Model} & \textbf{Accuracy} & \textbf{Search Calls} & \textbf{Get Document Calls} & \textbf{Calibration Error} \\
      \midrule
      gpt-4.1 & 35.42\% & 8.67 & N/A & 54.67\% \\
      gpt-4.1 + get-doc & 43.61\% & 10.03 & 1.85 & 54.28\% \\
      \midrule
      Qwen3-32B & 10.36\% & 0.94 & N/A & 59.84\% \\
      Qwen3-32B + get-doc & 11.69\% & 1.01 & 0.27 & 56.47\% \\
      \bottomrule
    \end{tabular}
  \endgroup
\end{table}

\subsubsection{Effect of Document Reading Strategy}
\label{sec:getdoc}
In previous experiments, we always presented only the first 512 tokens of each retrieved document as a preview to the LLM during each round of search and reasoning, due to token budget constraints. However, in realistic deep research scenarios, agents often have access to a document reader tool that enables reading the full content of a document. To evaluate the potential benefit of such a tool, we conduct experiments with gpt-4.1 and Qwen3-32B, both with and without access to a whole-document reader (referred to as the get-document tool). Appendix~\ref{appendix:get_doc_prompt} contains the revised prompt used when the get-document tool is added.

Results are shown in Table~\ref{table:get_doc}. For gpt-4.1, enabling the get-document tool improves answer accuracy from 35.42\% to 43.61\%, with a modest increase in search calls (from 8.67 to 10.03) and an average of 1.85 full-document reads per query. This confirms that having access to full documents provides additional useful context that enhances final decision-making. 

For Qwen3-32B, which performs worse overall, the benefit is more modest. Accuracy improves slightly from 10.36\% to 11.69\%, and the number of get-document calls remains low (0.27 per query on average). This suggests that while the tool can help, the model's limited reasoning and tool-use ability constrain its ability to exploit the additional information. 

These results show that the whole-document reading tool can improve performance, especially for strong models like gpt-4.1, by providing access to richer context beyond truncated previews.
However, its effectiveness depends heavily on the agent's capability to recognize when and how to use the tool, highlighting once again the importance of model quality in effective tool integration. 
This also highlights the value of context engineering in optimizing how retrieval results are presented to the LLM agent.

\begin{table}[]
  \centering
  \caption{Evidence document retrieval effectiveness on the Fineweb 10BT corpus.}
  \label{table:fineweb_retrieval}
  \begingroup
    \tableformat
    \begin{tabular}{llcccc}
      \toprule
      \textbf{Retriever} & \textbf{Corpus} & \textbf{Recall@5} & \textbf{Recall@100} & \textbf{Recall@1000} & \textbf{nDCG@10} \\
      \midrule
      BM25        & Original     & 1.2\% & 4.7\% & 13.6\% & 1.6\% \\
      BM25  &  Original + Fineweb   & 2.2\% & 8.0\% & 19.4\% & 3.1\% \\
      \midrule
      Qwen3-Embed-8B  & Original         & 14.5\% & 47.7\% & 76.7\% & 20.3\% \\
      Qwen3-Embed-8B &  Original + Fineweb & 11.6\% & 37.6\% & 64.2\% & 16.4\% \\
      \midrule
      ReasonIR-8B   & Original          & 12.2\% & 43.6\% & 73.9\% & 16.8\% \\
      ReasonIR-8B &  Original + Fineweb    & 8.6\% & 30.7\% & 56.3\% & 11.8\% \\
      \bottomrule
    \end{tabular}
  \endgroup
\end{table}

\begin{table}[]
  \centering
\caption{Accuracy of end-to-end search agents on our \model\ original 100k corpus vs. FineWeb 10BT corpus.}

  \label{table:fineweb-accuracy-corpus}
  \resizebox{0.7\linewidth}{!}{%
  \begin{tabular}{lll r}
    \toprule
    \textbf{LLM} & \textbf{Retriever} & \textbf{Corpus} & \textbf{Accuracy} \\
    \midrule
    \multirow{4}{*}{SearchR1-32B}
      & BM25                          & Original             & 3.86\% \\
      & BM25                          & Original + Fineweb   & 4.72\% \\
      & Qwen3-Embed-8B                & Original             & 10.36\% \\
      & Qwen3-Embed-8B                & Original + Fineweb   & 8.33\% \\
    \midrule
    \multirow{4}{*}{Qwen3-32B}
      & BM25                          & Original             & 3.49\% \\
      & BM25                          & Original + Fineweb   & 5.42\% \\
      & Qwen3-Embed-8B                & Original             & 10.36\% \\
      & Qwen3-Embed-8B                & Original + Fineweb   & 7.11\% \\
    \bottomrule
  \end{tabular}%
  }
\end{table}

\subsubsection{Effect of Corpus Size}

The corpus in \model contains approximately 100K documents. While real-world agents often operate over much larger, web-scale corpora, we aim to assess whether our designed corpus size is sufficient to support valid experimental observations. To this end, we augment our benchmark corpus with the Fineweb-edu~\cite{penedo2024the} document collection (10 billion tokens)\footnote{\url{https://huggingface.co/datasets/HuggingFaceFW/fineweb-edu/viewer/sample-10BT}}, deduplicated by URL. This expansion results in a significantly larger corpus of 9,771,311 documents-roughly 10 times larger than the original.


Table~\ref{table:fineweb_retrieval} shows retrieval performance before and after adding Fineweb documents. For BM25, retrieval effectiveness improves across all metrics, likely due to better inverse document frequency (IDF) estimation in the larger corpus, which strengthens BM25’s lexical scoring.

In contrast, neural retrievers (Qwen3-Embedding-8B and ReasonIR-8B) show degraded performance on the Fineweb-augmented corpus. This drop is theoretically expected: the relative ranking of documents from the original small corpus remains unchanged, but the newly added Fineweb documents can now appear in the top ranks. Since these additional documents are unjudged, they are treated as non-relevant under standard TREC-style evaluation, inevitably lowering measured retrieval effectiveness.

It is important to note that lower retrieval scores for embedding models on Fineweb do not necessarily indicate worse final answers, some unjudged, top-ranked Fineweb documents may be ``false negatives'' that still provide useful evidence. However, as shown in Table~\ref{table:fineweb-accuracy-corpus}, adding Fineweb does not improve answer accuracy for embedding-based retrievers. For example, Qwen3-32B with Qwen3-Embedding-8B drops from 10.36\% to 7.11\% accuracy.

Overall, expanding the corpus size by a factor of 10 does not lead to different conclusions about the ranking or effectiveness level among the retrievers and LLM search agents, supporting our claim that the original 100K corpus offers both strong positive coverage and sufficient challenge for robust evaluation.

\section{Future Work and Discussion}
We believe that our \model opens new avenues for advancing research in the Deep-Research area. 
\model retains the challenging nature of the original BrowseComp while providing a more controlled and transparent experimental setup similar to early pivotal evaluation benchmarks like Natural Question (NQ)~\cite{kwiatkowski-etal-2019-natural} and HotpotQA~\cite{yang-etal-2018-hotpotqa}.
Like how NQ and HotpotQA have facilitated the design, comparison, and diagnosis of modern neural QA systems, we hope that \model will serve similar roles for Deep-Research agent studies. Here, we list some immediate research directions.

While our current work focuses on how different retrievers influence inference performance, a promising future direction is to examine the role of the retriever during agent optimization. For example, optimizing a search agent may be more challenging when paired with BM25 than with a modern embedding-based retriever, simply because BM25 surfaces fewer relevant documents. Understanding how retriever quality affects the learning dynamics of an agent remains an open question.

Another important extension is to study the agent’s `out-of-distribution' tool-use capabilities. For instance, if an agent is optimized using a BM25 search tool, how well does its performance generalize when switched to an embedding-based search tool?

A more creative research could be an attempt on a breakdown of the commercial search engine. As much as a folktale, a commercial search solution employs tiered, composed, and multi-facet search solution. Is the LLM able to orchestrate a set of search tools to perform federated search~\cite{fedsearch}, or even a sub-agent, to get quality results similar to those from Google?

A further direction is to design retrieval models that are tolerant of, or even adaptive to, a specific agent. In the Deep Research setting, the primary consumer of retrieved documents is no longer a human, but a tool-augmented LLM agent. This raises the possibility that retrieval models could be co-optimized with the agent for achieving overall answer accuracy, rather than developed and evaluated in isolation.

Finally, as shown in this work, an oracle retriever capable of surfacing gold or highly relevant documents can greatly improve accuracy. Such retrievers may also reduce the number of search iterations required, improving the overall efficiency of the research process. Developing high-precision retrieval systems for reasoning-intensive, complex queries could yield substantial benefits for real-world applications.

Overall, \model serves as an ideal testbed for pursuing these directions, enabling systematic and fine-grained analyses of agent–retriever interactions within the Deep-Research paradigm.

\section{Conclusion}

We introduced \model, a new benchmark designed to address the reproducibility, fairness, and transparency challenges in evaluating Deep-Research Agents. By grounding each query in a fixed, human-verified corpus containing both positive and hard-negative documents, our framework enables the independent and controlled assessment of retrieval and agent components.

Through extensive experiments pairing diverse retrievers with both open- and closed-source agents, we demonstrate that retrieval quality substantially impacts both the effectiveness and efficiency of deep research systems. Stronger retrievers not only improve final answer accuracy but also reduce the number of search iterations required, while oracle-level retrieval reveals the significant headroom still available for progress.

\model provides a robust platform for probing these dynamics and paves the way for future research on co-optimizing retrievers and agents, improving out-of-distribution tool-use generalization, and advancing context engineering frameworks. By making our benchmark and baselines publicly available, we aim to catalyze the next generation of Deep-Research systems.

\section*{Acknowledgment}
We extend our sincere thanks to Guido Zuccon, Bevan Koopman, Xin Zhang for their valuable and insightful discussions.

\bibliography{neurips_2024}
\bibliographystyle{unsrtnat}

\clearpage

\appendix

\section{OpenAI O3 Evidence Document Gathering Prompt}\label{appendix:gather_doc}

\begin{tcolorbox}
I will give you a question and a correct answer, and you are to search online for evidence that supports the answer. List the evidence you've used to justify this answer step-by-step, including their urls in your output. Your final list of urls should be in the order such that a human can visit them in order to justify the answer. \\

Question: \{question\}\\

Answer: \{answer\}\\

This is all the information you have to work with to produce the final list of urls. Format your answer in a table with 3 columns: \\
- clue: the clue mentioned in the question \\
- url: the http web url of the evidence you've found \\
- evidence: the content in the url page that supports the clue

\end{tcolorbox}

\section{Labelling UI Example}\label{appendix:ui}

\begin{figure}[H]
    \centering
    \includegraphics[width=\columnwidth]{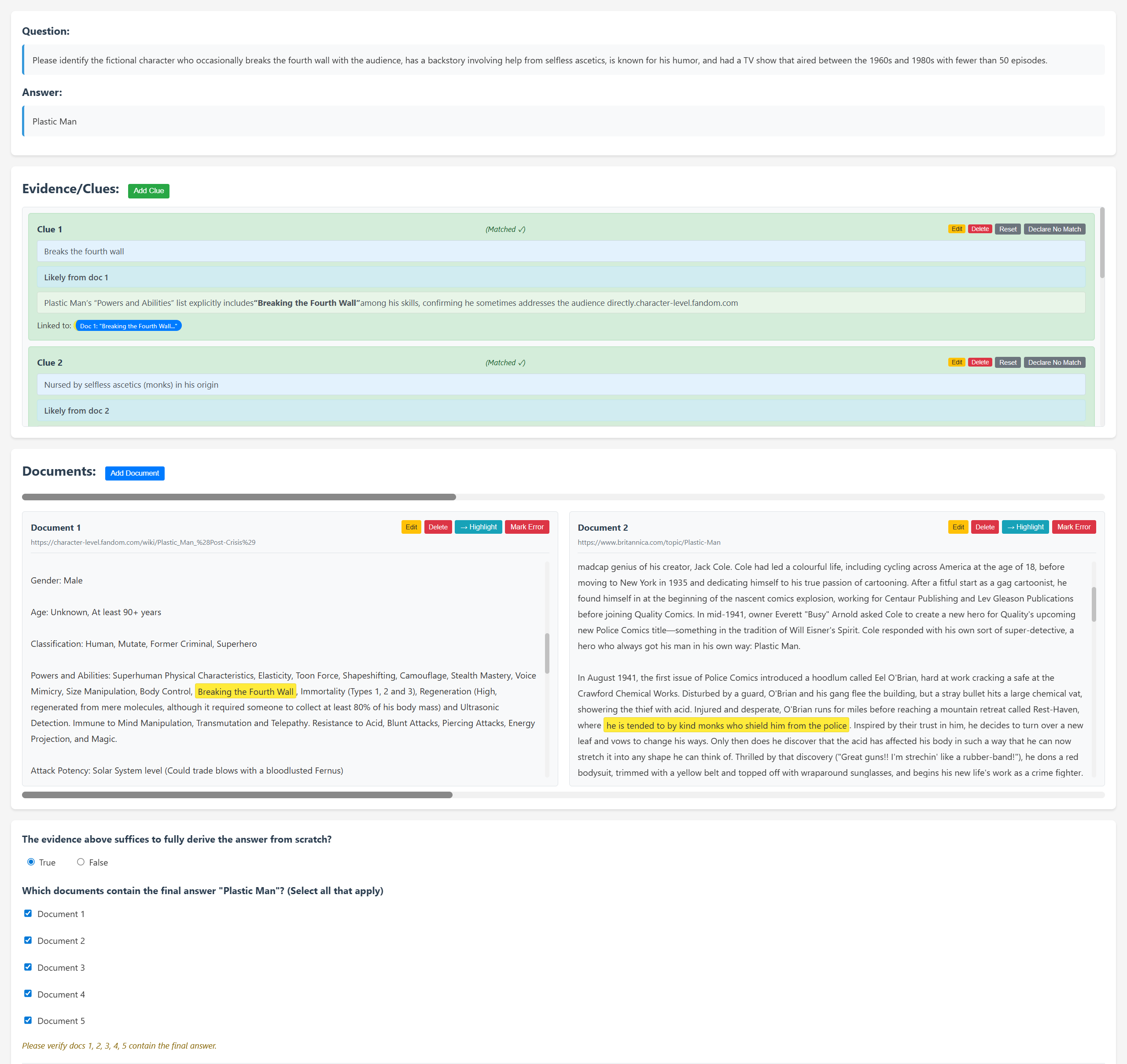}
    \caption{A screenshot of the annotation interface.}
\end{figure}

\section{Problematic Cases}
\label{sec:problematic}
\begin{itemize}
    \item BrowseComp Errors: During the verification process, we discover that some question-answer pairs in BrowseComp are inherently flawed. For example, one question asks for the name of a book whose author later returned to acting. Using the ground-truth answer, we can identify the intended book and its listed author. However, upon further investigation, we find that the individual who wrote the book and the one who returned to acting are two different people who happen to share the same name.

    \item Extensive Use of Google Maps: 42 queries in BrowseComp require distance-related information that explicitly prompt multiple calls to Google Maps. These are removed because high-quality documents discussing specific Google Maps distances between arbitrary locations are difficult to obtain. Moreover, scraping static snapshots of Google Maps pages to include in the corpus is not a valid substitute; answering such questions as intended should require agents to be augmented with access to the Google Maps API, rather retrieving from a corpus. However, this capability lies outside the scope of our objective to build a static, document-based dataset.

    \item Ambiguous or Non-Unique Answers: Some question-answer pairs are well-supported by documents, but suffer from ambiguity in the expected answer format or the existence of multiple valid answers. For instance, one question asks for the username of an individual who authored a specific story on an internet forum. While the ground-truth answer is correct, it is only one of three usernames credited as authors. We remove 13 such queries due to this kind of ambiguity.
\end{itemize}

\section{Negative Mining Query Decomposition Prompt}\label{appendix:decompose_prompt}

\begin{tcolorbox}
You are an expert at breaking down complex, multi-part questions into simpler, self-contained subqueries.  \\

Your task is to analyze the given question and decompose it into a series of smaller, more manageable subqueries that, when answered together, would provide all the information needed to answer the original question. \\

Guidelines: \\

1. Each subquery should focus on a single piece of information or concept

2. Subqueries MUST be completely self-contained and answerable independently - do not use pronouns or references like "this person", "the author", "these conditions", "they", "the movie", etc.

3. Each subquery should include all necessary context and constraints from the original query

4. Preserve all important details and constraints from the original query

5. Return only the subqueries as a JSON array of strings \\

Example: \\

Original: "Please identify the fictional character who occasionally breaks the fourth wall with the audience, has a backstory involving help from selfless ascetics, is known for his humor, and had a TV show that aired between the 1960s and 1980s with fewer than 50 episodes." \\

Subqueries:
[
"Which fictional characters occasionally break the fourth wall with the audience?", 
"Which fictional characters have a backstory involving help from selfless ascetics?",
"Which fictional characters are known for their humor?",
"Which TV shows aired between the 1960s and 1980s?",
"Which TV shows had fewer than 50 episodes?
] \\

Please decompose this query into subqueries:

\{query\}

\end{tcolorbox}

\section{Main Search Prompt}\label{appendix:main_prompt}

\begin{tcolorbox}

You are a deep research agent. You need to answer the given question by interacting with a search engine, using the search tool provided. Please perform reasoning and use the tool step by step, in an interleaved manner. You may use the search tool multiple times. \\

Question: \{Question\} \\

Your response should be in the following format:

Explanation: \{\{your explanation for your final answer. For this explanation section only, you should cite your evidence documents inline by enclosing their docids in square brackets [] at the end of sentences. For example, [20].\}\}

Exact Answer: \{\{your succinct, final answer\}\}

Confidence: \{\{your confidence score between 0\% and 100\% for your answer\}\}

\end{tcolorbox}

\section{Evaluation Prompt}\label{appendix:eval_prompt}

\begin{tcolorbox}

Judge whether the following [response] to [question] is correct or not based on the precise and unambiguous [correct\_answer] below. \\

[question]: \{question\} \\

[response]: \{response\} \\

Your judgement must be in the format and criteria specified below: \\

extracted\_final\_answer: The final exact answer extracted from the [response]. Put the extracted answer as `None' if there is no exact, final answer to extract from the response. \\

[correct\_answer]: \{correct\_answer\} \\

reasoning: Explain why the extracted\_final\_answer is correct or incorrect based on [correct\_answer], focusing only on if there are meaningful differences between [correct\_answer] and the extracted\_final\_answer. Do not comment on any background to the problem, do not attempt to solve the problem, do not argue for any answer different than [correct\_answer], focus only on whether the answers match. \\

correct: Answer `yes' if extracted\_final\_answer matches the [correct\_answer] given above, or is within a small margin of error for numerical problems. Answer `no' otherwise, i.e. if there if there is any inconsistency, ambiguity, non-equivalency, or if the extracted answer is incorrect. \\

confidence: The extracted confidence score between 0|\%| and 100|\%| from [response]. Put 100 if there is no confidence score available.

\end{tcolorbox}

\section{Search Prompt with Get-Doc}\label{appendix:get_doc_prompt}

\begin{tcolorbox}

You are a deep research agent. You need to answer the given question by interacting with a search engine, using the search and get\_document tools provided. Please perform reasoning and use the tools step by step, in an interleaved manner. You may use the search and get\_document tools multiple times. \\

Question: \{Question\} \\

Your response should be in the following format: \\

Explanation: \{\{your explanation for your final answer. For this explanation section only, you should cite your evidence documents inline by enclosing their docids in square brackets [] at the end of sentences. For example, [20].\}\}

Exact Answer: \{\{your succinct, final answer\}\}

Confidence: \{\{your confidence score between 0\% and 100\% for your answer\}\}

\end{tcolorbox}

\section{API Cost}
\label{sec:api_cost}
\begin{table}[]
  \centering
  \caption{Overall API cost of proprietary agents.}
    \label{table:money}
  \begingroup
    \tableformat
    \begin{tabular}{llcc}
      \toprule
      \textbf{LLM} & \textbf{Retriever} & \textbf{Accuracy} & \textbf{Price (USD)} \\
      \midrule
      \multirow{2}{*}{gpt-4.1}
        & BM25                   & 14.58\% & \$106.96 \\
        & Qwen3-Embed-8B     & 35.42\% & \$89.81 \\
      \midrule
      \multirow{2}{*}{o3}
        & BM25                   & 49.28\% & \$836.35 \\
        & Qwen3-Embed-8B     & 63.49\% & \$740.79 \\
    \midrule
      \multirow{2}{*}{GPT-5}
        & BM25                   & 55.9\% & \$400.36 \\
        & Qwen3-Embed-8B     & 70.12\% & \$360.71 \\
      \midrule
      \multirow{2}{*}{Sonnet\,4}
        & BM25                   & 14.34\% & \$352.04 \\
        & Qwen3-Embed-8B     & 36.75\% & \$325.75 \\
      \midrule
      \multirow{2}{*}{Opus\,4}
        & BM25                   & 15.54\% & \$2,043.95 \\
        & Qwen3-Embed-8B     & 36.14\% & \$1,842.48 \\
      \midrule
      \multirow{2}{*}{Gemini~2.5~Flash}
        & BM25                   & 15.54\% & \$47.32 \\
        & Qwen3-Embed-8B     & 33.01\% & \$41.29 \\
      \midrule
      \multirow{2}{*}{Gemini~2.5~Pro}
        & BM25                   & 19.04\% & \$138.64 \\
        & Qwen3-Embed-8B     & 28.67\% & \$99.92 \\
      \bottomrule
    \end{tabular}
  \endgroup
\end{table}

Table~\ref{table:money} Shows the API cost of the experiments.

\newpage
\end{document}